\documentclass[journal]{IEEEtran}

\usepackage{amsmath,amssymb,amsfonts,amsthm}
\usepackage{mathtools}
\usepackage{bm}
\usepackage{booktabs}
\usepackage{multirow}
\usepackage{makecell}
\usepackage{array}
\usepackage{graphicx}
\usepackage{xcolor}
\usepackage{algorithm}
\usepackage{algpseudocode}
\usepackage{tikz}
\usepackage{svg}
\usepackage{graphicx}
\usepackage{comment}

\algrenewcommand\algorithmicrequire{\textbf{Input:}}
\algrenewcommand\algorithmicensure{\textbf{Output:}}
\algnewcommand{\Phase}[1]{\Statex\hspace*{1.2em}\(\triangleright\)~\textbf{#1}}

\usetikzlibrary{arrows.meta,positioning,shapes.geometric,calc,fit,backgrounds,shadows.blur}
\usepackage{pgfplots}
\pgfplotsset{compat=1.18}
\usepgfplotslibrary{fillbetween}
\usetikzlibrary{patterns}
\usepackage{cite}
\usepackage{pifont}
\usepackage[hidelinks]{hyperref}
\usepackage{url}
\usepackage{enumitem}

\newcommand{\Scal}{\mathcal{S}}

\definecolor{qctgreen}{RGB}{15,110,86}
\definecolor{qctteal}{RGB}{29,158,117}
\definecolor{qctpurple}{RGB}{83,74,183}
\definecolor{qctgray}{RGB}{136,135,128}
\definecolor{qctblue}{RGB}{55,138,221}
\definecolor{qctcoral}{RGB}{216,90,48}
\definecolor{qctamber}{RGB}{239,159,39}
\definecolor{qctred}{RGB}{163,45,45}

\usepackage[table]{xcolor}

\definecolor{bestblue}{HTML}{A6CEE3}     
\definecolor{secondblue}{HTML}{DEEBF7}   
\newcommand{\best}[1]{\cellcolor{bestblue}\textbf{#1}}
\newcommand{\second}[1]{\cellcolor{secondblue}\underline{#1}}

\tikzset{
  qnode/.style={rounded corners=4pt, draw=#1, fill=#1!10, line width=0.4pt,
                inner sep=3pt, font=\footnotesize, align=center},
  qarrow/.style={-Latex, line width=0.5pt, draw=black!60},
  qdash/.style={-Latex, line width=0.4pt, draw=black!50, dashed},
}

\begin{document}

\title{CaRe-KT: Continuity-Aware Retrieval-Augmented Reasoning for Knowledge Tracing}
\author{Mehrnoush~Mohammadi,
        Kamal~Berahmand, Mohammad Morsali, Shazia Sadiq, Wojtek Tomaszewski,
        and~Hassan~Khosravi%

\thanks{M. Mohammadi and S. Sadiq are with the School of Electrical Engineering and Computer Science, The University of Queensland, Brisbane, Australia (e-mail: m.mohammadibolbanabad@uq.edu.au; shazia@eecs.uq.edu.au).}%
\thanks{K. Berahmand is with the School of Computing Technologies, RMIT University, Melbourne, Australia (e-mail: kamal.berahmand@rmit.edu.au).}%
\thanks{M. Morsali is with the School of Telecommunications and Aerospace Engineering , Universitat Polit`ecnica de Catalunya, Spain (e-mail: mohammadmorsali1381@gmail.com).}
\thanks{W. Tomaszewski is with the Australian Research Council Centre of Excellence for Children and Families over the Life Course, The University of Queensland, Brisbane, Australia (e-mail: w.tomaszewski@uq.edu.au.)}%
\thanks{H. Khosravi is with the Institute for Teaching and Learning Innovation, The University of Queensland, Brisbane, Australia (e-mail: h.khosravi@uq.edu.au).}%

\thanks{Corresponding author: Mehrnoush Mohammadi.}}

\markboth{Preprint submitted to IEEE Transactions on Learning Technologies}%
{Mohammadi \MakeLowercase{\textit{et al.}}: Continuity-Aware Retrieval-Augmented Reasoning for Knowledge Tracing}
\IEEEaftertitletext{%
\parbox{\textwidth}{%
\centering\normalfont\mdseries\upshape\footnotesize
This work has been submitted to the IEEE for possible
publication. Copyright may be transferred without notice,
after which this version may no longer be accessible.
}%
\vspace{0.1 cm}
}

\maketitle
\begin{abstract}
Knowledge tracing (KT) estimates learners’ evolving knowledge states from their interaction histories and underpins exercise recommendation and instructional support in adaptive learning systems. Recent retrieval-augmented KT models improve these estimates by explicitly accessing cross-learner behavioral evidence at prediction time. However, most follow a retrieve-then-predict pipeline in which relevance governs evidence acquisition, while the predictive influence of retrieved evidence is determined separately during downstream inference. This decoupling leaves a retrieval–inference alignment gap in KT, as behavioral relevance varies with the learner’s evolving state and prediction context, potentially undermining reliable knowledge estimation and well-targeted adaptive learning support. To address this gap, we propose CaRe-KT, a query-adaptive framework that maintains relevance continuity between evidence acquisition and inference without equating retrieval relevance with predictive utility. CaRe-KT introduces an Evidence-Graph Cache (EGC) that combines persistent question-level relational compatibility with interaction-level temporal validity, progressively refining query-relevant questions into temporally valid historical anchors that form a query-specific evidence subgraph. Each anchor retrieves the learning trajectory that preceded it, and these trajectories are encoded with the target learner’s history in a shared representation space shaped by outcome-supervised contrastive learning. CaRe-KT then reasons over this evidence through attention that combines the carried acquisition relevance with learned query–evidence compatibility, so that relevance informs the influence of each trajectory without determining it. Across five benchmarks, CaRe-KT outperforms strong KT baselines and remains robust under sparse-history and cold-start conditions, supporting relevance continuity as a design principle for retrieval-augmented KT.
\end{abstract}

\begin{IEEEkeywords}
Knowledge Tracing, Retrieval-Augmented Reasoning, Query-Adaptive Retrieval, Relevance Continuity. 
\end{IEEEkeywords}

\section{Introduction}
\begin{figure}
    \centering
    \includegraphics[width=1\linewidth]{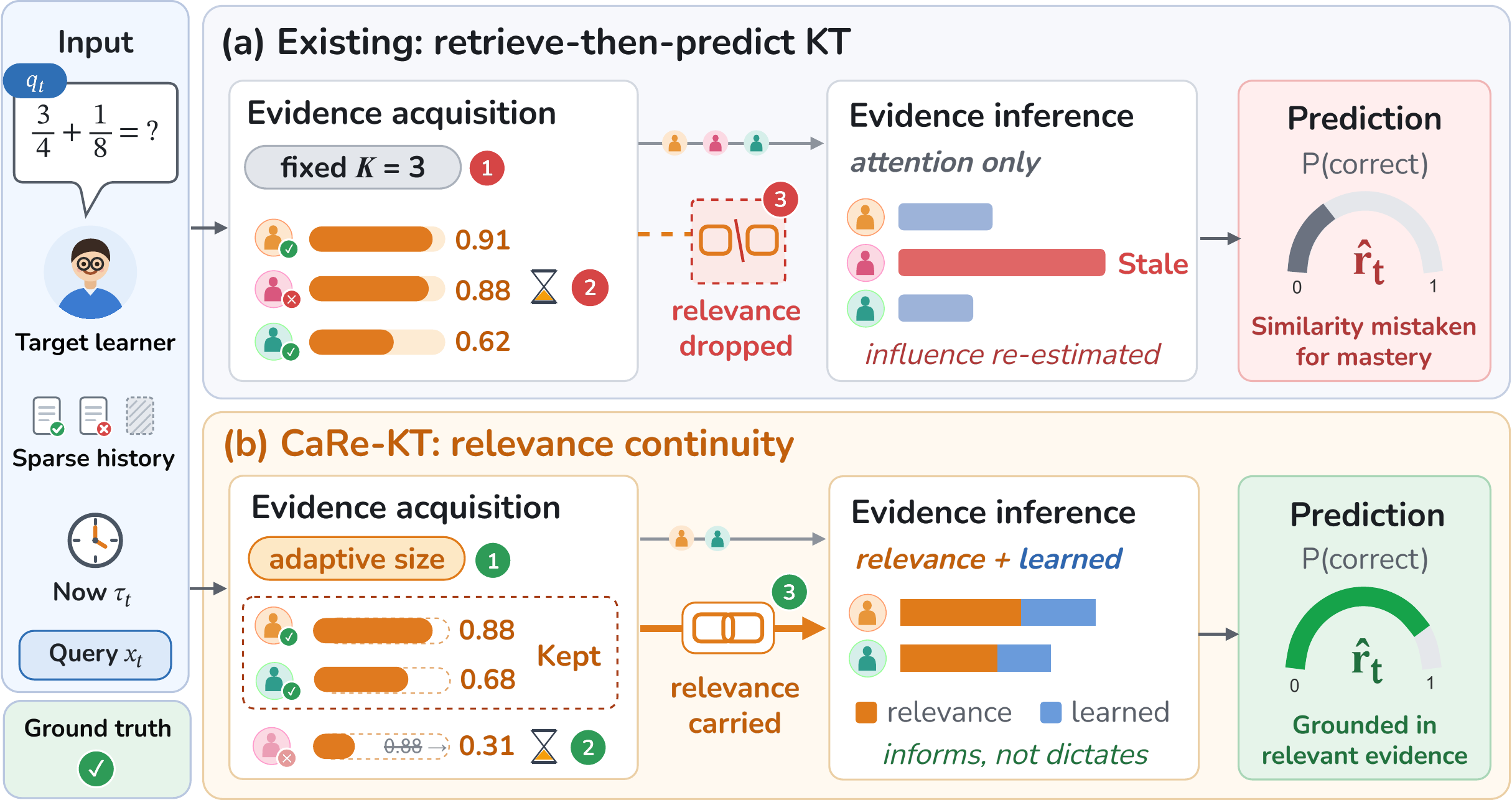}
\caption{Retrieval--inference alignment in retrieval-augmented KT. (a) Existing pipelines admit a fixed number of evidence items \textcircled{1} by time-agnostic relevance \textcircled{2} and discard that relevance before inference \textcircled{3}, so stale evidence can dominate. (b) CaRe-KT admits a query-adaptive evidence set \textcircled{1}, refines relevance by temporal validity \textcircled{2}, and carries it into inference \textcircled{3}, where it informs but does not dictate influence. Values are illustrative.}
    \label{fig:introGap}
\end{figure}

\IEEEPARstart{K}{nowledge} tracing (KT) aims to infer a learner's evolving knowledge state from historical interactions, typically through the prediction of future performance. Reliable estimation of this latent state is central to intelligent tutoring and adaptive learning systems, where it informs decisions such as personalized exercise recommendation, learning-path adaptation, and the timely provision of instructional support. Over the past decade, KT has progressed from probabilistic latent-state models to increasingly expressive neural architectures based on recurrent networks~\cite{piech2015deep}, memory networks~\cite{mohammadi2026augmenting}, attention mechanisms~\cite{hou2025cognitive, deng2025denoised}, and graph neural networks~\cite{cheng2024dygkt, li2025hyperbolic}. These models have substantially improved predictive performance by learning complex dependencies within a student's interaction sequence~\cite{shen2024survey}.

Despite these advances, most conventional KT models remain fundamentally \emph{history-centric}, estimating a learner's current state primarily from its own observed trajectory under globally shared parameters learned from population-level interaction data. Cross-learner behavioral regularities are therefore implicitly absorbed within these shared parameters, with no explicit mechanism for accessing specific historical experiences that may be informative for the current learner state~\cite{long2022improving, duan2026hise}. This limitation becomes particularly restrictive when a learner's history is sparse, noisy, or unreliable, as is common for new learners with few recorded interactions, low-activity learners with fragmented practice, and newly introduced questions with little response history~\cite{lu2022cmkt, zhang2024coskt}.

Recent collaborative and retrieval-enhanced KT methods have begun to relax this parametric paradigm by externalizing historical interaction records as retrievable evidence and selectively accessing relevant cross-learner behavioral information at prediction time~\cite{long2022improving,duan2026rag,duan2026microc}. This shift introduces an explicit evidence acquisition layer into knowledge-state inference, making prediction dependent not only on encoding the target learner's trajectory, but also on establishing which external behavioral evidence is relevant to the current prediction context. Existing methods typically realize this process through a retrieve-then-predict pipeline, in which relevant historical interactions are first identified by an explicit relevance criterion and then encoded and aggregated for prediction. However, this decomposition introduces a retrieval–inference alignment gap, where relevance governs which historical interactions are admitted as evidence, while their predictive influence is determined separately during downstream inference without explicitly preserving the relevance relation established during retrieval~\cite{zhou2026optimizing, xu2025training, lin2024ra, zhang2025utility, tao2026task}(Fig.~\ref{fig:introGap}(a)). Educationally, this disconnect can undermine both the reliability and accountability of learner-state estimation. When the influence of cross-learner evidence becomes detached from its relevance to the target learner, the system risks mistaking behavioral similarity for evidence of mastery, leading to instructional decisions that reinforce irrelevant learning patterns, overlook real knowledge gaps, or misjudge readiness for progression. Furthermore, such decisions may lack a defensible evidential basis, limiting educators' ability to scrutinize and justify interventions that affect learners' progress.

Closing this gap, however, requires more than carrying retrieval scores into downstream prediction. Once retrieval-derived relevance remains active during reasoning, it becomes part of the context through which retrieved evidence is interpreted. A coarse or static relevance relation may therefore be insufficient, since the usefulness of historical behavior can vary across prediction contexts even when the evidence itself appears related to the query~\cite{tian2025relevance, zhao2026r}. This is particularly consequential in KT, where behavioral relevance is inherently non stationary. As the learner state evolves, both the informativeness of a given historical interaction and the amount of evidence worth using change from one query to another. Yet the relevance available in current pipelines is usually established over a predefined number of candidates and without explicitly accounting for the temporal validity of individual interactions~\cite{long2022improving, zhang2024coskt, duan2026rag, duan2026microc}, so carrying it forward would let outdated or weakly related interactions shape inference for some queries while informative ones remain excluded for others. Behavioral relevance should therefore retain persistent relational information while adapting its interpretation to the evolving prediction context.

Motivated by this principle, we propose CaRe-KT (Fig.~\ref{fig:introGap}(b)), a query-adaptive retrieval-augmented KT framework that realizes relevance continuity through a shared relevance structure. CaRe-KT acquires evidence through an Evidence-Graph Cache (EGC), which combines persistent question-level relational compatibility with interaction-level temporal validity to form, for each query, an evidence subgraph of temporally valid historical interactions whose edge weights record acquisition relevance. Each selected interaction anchors a learning trajectory that preceded it, and these trajectories are encoded alongside the target learner's history in a shared behavioral representation space, shaped by outcome-supervised contrastive learning and query-consistency regularization. CaRe-KT then reasons over the learned representations through relevance-informed attention, where evidence-subgraph relevance is combined with learned query-specific compatibility, allowing retrieval relevance to guide the influence of retrieved evidence without directly determining its predictive contribution. Extensive experiments on five benchmark datasets demonstrate that CaRe-KT consistently outperforms strong KT baselines and remains robust under sparse and cold-start conditions.

The key contributions of this work are as follows.
\begin{itemize}
\item We identify the retrieval-inference alignment gap in retrieval-augmented KT and introduce relevance continuity as a design principle that preserves the rationale for evidence acquisition throughout reasoning without equating retrieval relevance with predictive utility.
\item We propose the EGC, a query-adaptive evidence acquisition mechanism that refines persistent question-level relations into temporally valid, query-specific evidence subgraph with adaptive cardinality, enabling trajectory-grounded retrieval under evolving prediction contexts.
\item We design a relevance-informed reasoning mechanism that couples carried acquisition relevance, as a trained prior, with learned query-evidence compatibility in attention, over a structured representation space shaped by
supervised contrastive and query-consistency regularization, so that relevance informs the influence of each retrieved trajectory while compatibility adapts it to the current query.
\item We evaluate CaRe-KT on five benchmark datasets against thirteen baselines, demonstrating consistent improvements in predictive performance, robustness under sparse-history and question cold-start conditions, and the effectiveness of relevance continuity through ablation studies.
\end{itemize}
\section{Related Work}
\subsection{Parametric Knowledge Tracing}
Knowledge tracing (KT) has evolved from probabilistic latent-state models, such as Bayesian KT (BKT), to neural architectures that capture increasingly complex temporal and structural dependencies in learners' interactions~\cite{vsaric2024twenty,shen2024survey}. 

DKT~\cite{piech2015deep} models knowledge-state evolution through a recurrent hidden state. Building on this formulation, IEKT~\cite{long2021tracing}, DIMKT~\cite{shen2022assessing}, and QIKT~\cite{chen2023improving} introduce individual cognition, question difficulty, and question-centric representations, respectively. In contrast to a single recurrent state, memory-augmented approaches~\cite{zhang2017dynamic,zhang2024explore} maintain a separate knowledge representation for each concept. Attention-based models, including SAKT~\cite{pandey2019sakt}, AKT~\cite{ghosh2020context}, SparseKT~\cite{huang2023sparsekt}, simpleKT~\cite{liu2023simplekt}, stableKT~\cite{li2024enhancing}, and UKT~\cite{cheng2025uncertainty}, selectively capture dependencies among historical interactions. Graph-based approaches further incorporate relational structure through concept graphs~\cite{nakagawa2019gkt}, question--concept associations~\cite{yang2020gikt}, and dynamic student-question interactions in DyGKT~\cite{cheng2024dygkt}. Recent hybrid architectures integrate complementary modeling mechanisms to improve the characterization of evolving knowledge states~~\cite{he2023man,liu2026enhancing,mohammadi2026augmenting}

Despite these diverse architectures, knowledge-state estimation remains primarily grounded in the target learner’s observed trajectory, with broader behavioral regularities primarily encoded in shared model parameters and learned representations rather than explicitly accessed as query-relevant instance-level evidence at inference time~\cite{long2022improving}. This reliance becomes particularly restrictive under the sparse and highly skewed interaction patterns common in educational data, where many questions are sparsely observed and individual learners may have insufficient interaction histories for reliable knowledge-state estimation~\cite{pu2024elakt, ma2025learning}.

\subsection{Retrieval-Augmented Knowledge Tracing}
Collaborative KT methods extend individual-history inference by explicitly incorporating behavioral evidence from other learners. CoKT~\cite{long2022improving} retrieves peer records that share the target question or its concepts, ranks them by BM25 similarity of their preceding sequences, and combines the attention-weighted evidence with the learner's own representation for prediction. CoSKT~\cite{zhang2024coskt} retrieves students whose learning experiences overlap with the target learner, derives a common knowledge state from their sequences, and fuses it with the individual state through gating under a self-supervised consistency constraint. RACL~\cite{berahmand2026retrieval} selects historical interaction prefixes by pedagogical similarity over representation, concept overlap, and difficulty, and uses them both as inter-learner contrastive supervision and as evidence aggregated through target-conditioned cross-attention. FINER~\cite{liu2025advancing} shifts the retrieval unit from peers to learning patterns, retrieving the follow-up performance trends of learners who exhibited a similar recent pattern and aggregating them through similarity-aware attention that weights each trend by its frequency and contextual similarity before combining it with the learner's own sequence.

These methods share a common design in which retrieved evidence is weighted by a learned aggregator. A second group instead structures the search space for evidence acquisition and delegates prediction to a large language model. RAG-KT~\cite{duan2026rag} aligns multi-source student information through question groups, retrieves similar students by fusing behavioral, structural, and ability-related similarities, and assembles the retrieved evidence as structured context for LLM-based prediction. MicroC-KT~\cite{duan2026microc} organizes learners into latent communities, matches the target learner to a community using bi-granular summaries, and retrieves peer evidence within that community through multi-route ranking before structured prompting.

Collectively, these methods demonstrate the value of explicit cross-learner evidence, yet largely retain a \emph{retrieve-then-predict} decomposition in which relevance determines evidence acquisition while downstream influence is reconstructed separately through attention, gating, or prompting. Beyond this separation, evidence acquisition itself is often only partially adaptive: predefined top-\(K\) budgets impose a uniform evidence cardinality across heterogeneous queries, while relevance is typically established over peer-, sequence-, or community-level candidates without explicitly refining individual interactions by their temporal validity for the current query.
\subsection{Retrieval-Inference Alignment in Retrieval-Augmented Learning}
Beyond KT, recent retrieval-augmented reasoning increasingly questions a strict separation between evidence acquisition and downstream inference~\cite{trivedi2023interleaving}. R$^2$AG~\cite{ye2406r2ag} identifies a semantic gap between retrievers and generators arising from their distinct objectives and shows that propagating retrieval information into generation better anchors downstream reasoning, while ARL2~\cite{zhang2024arl2} similarly demonstrates the benefits of aligning retrieval with downstream model feedback. Alignment alone does not settle how retrieved evidence should be used, however, as recent empirical studies show that relevance, although informative, does not by itself determine the utility of retrieved evidence~\cite{salemi2024evaluating,tian2025relevance,zhang2025utility}. This finding has motivated methods that keep the influence of retrieved evidence adaptive. Self-RAG~\cite{asai2024self} decides during generation whether and how retrieved content is used, REPLUG~\cite{shi2024replug} trains the retriever on language-model feedback, and RA-DIT~\cite{lin2024ra} adapts retriever and generator jointly. Taken together, these studies indicate that effective retrieval-augmented reasoning requires both \emph{relevance continuity} and sufficient flexibility for downstream utility to be adaptively determined during inference.

In KT, this balance is particularly consequential because behavioral relevance is both query- and time-dependent, while the downstream utility of retrieved evidence additionally depends on the learner's evolving state. Accordingly, retrieval relevance should inform, rather than dictate, downstream evidence use, making retrieval-inference alignment a central challenge under evolving learner contexts.
%
%
\section{Problem Formulation}
\begin{table}[!t]
\caption{Summary of Main Notation.}
\label{tab:notation}
\centering
\footnotesize
\renewcommand{\arraystretch}{1.08}
\setlength{\tabcolsep}{3pt}
\scalebox{0.91}{\begin{tabular}{ll}
\toprule
\textbf{Symbol} & \textbf{Description} \\
\midrule
$\mathcal{S},\mathcal{Q},\mathcal{C}$
& Student, question, and concept sets \\

$x_t=(s_t,q_t,\tau_t),\,r_t$
& Query interaction and response outcome \\

$\mathcal{C}_q,\,\mathbf{e}^{sem}_q$
& Concept set and semantic representation of $q$ \\

$\mathcal{H}^{s}_{<\tau_t}$
& Chronological history of student $s$ before $\tau_t$ \\

$\mathcal{M}_{<\tau_t}$
& Global interaction memory before $\tau_t$ \\

$\mathcal{R}_t,\,\Gamma_t$
& Retrieved behavioral evidence and its relevance structure \\

$ G^{t}=(\mathcal{V}^{t},\mathcal{E}^{t},\mathcal{W}^{t})$
& Active Evidence-Graph Cache (EGC) \\

$\mathrm{Obs}(q_i)$
& Historical occurrence buffer of question $q_i$ \\

$w_{ij},\,\tilde{w}_{ti}(\tau_k)$
& Relational compatibility and interaction-level relevance \\

$\epsilon_t,\,\mathcal{N}_{\epsilon}(q_t)$
& Adaptive relevance boundary and candidate question set \\

$\mathcal{K}_t$
& Selected historical interaction anchors \\

$\mathcal{G}^{E}_t = \left( \mathcal{V}^{E}_t, \mathcal{E}^{E}_t, \mathcal{W}^{E}_t \right)$
& Query-specific weighted evidence subgraph \\

$\mathcal{S}_i$
& Behavioral trajectory preceding anchor $i$ \\

$\mathbf{z}_t,\,\mathbf{z}_i$
& Query and evidence trajectory representations \\

$e_{ti},\,\alpha_{ti}$
& Query--evidence compatibility and attention weight \\

$\mathbf{h}_t,\,\hat r_t$
& Evidence-conditioned representation and prediction \\
\bottomrule
\end{tabular}
}
\end{table}
Let \(\mathcal{S}\), \(\mathcal{Q}\), and \(\mathcal{C}\) denote the sets of students, questions, and knowledge concepts, respectively, where each question \(q \in \mathcal{Q}\) is associated with a concept set \(\mathcal{C}_q \subseteq \mathcal{C}\). A student–question interaction is represented as \((s_t,q_t,\tau_t,r_t)\), where \(s_t \in \mathcal{S}\) denotes the student, \(q_t \in \mathcal{Q}\) the question, \(r_t \in \{0,1\}\) the response correctness, and \(\tau_t\) the interaction's position in the globally ordered event sequence. Table~\ref{tab:notation} summarizes the main notations used throughout the paper.

Given an incoming query $x_t=(s_t,q_t,\tau_t)$, conventional KT models estimate the probability that student $s_t$ correctly answers question $q_t$ at time $\tau_t$ as:
\begin{equation}
\hat{r}_t
=
P_{\Theta}
\left(
r_t=1
\mid
x_t,\mathcal{H}^{s_t}_{<\tau_t}
\right),
\end{equation}
where
$\mathcal{H}^{s_t}_{<\tau_t}
=
\{(q_j,\tau_j,r_j)\mid s_j=s_t,\ \tau_j<\tau_t\}$
denotes the chronological interaction history of student $s_t$ preceding $\tau_t$.

Beyond individual trajectories, let
$\mathcal{M}_{<\tau_t}
=
\{\mathcal{H}^{s}_{<\tau_t}\}_{s\in\Scal}$
denote the global behavioral memory. 
For retrieval-augmented KT, evidence acquisition is formulated as
\begin{equation}
\left(
\mathcal{R}_t,\Gamma_t
\right)
=
\Phi
\left(
x_t;\mathcal{M}_{<\tau_t}
\right),
\end{equation}
where $\Phi(\cdot)$ denotes the query-conditioned retrieval operator,
$\mathcal{R}_t$ is the resulting behavioral evidence, and $\Gamma_t$ denotes the relevance relations governing its selection.
The prediction is then formulated as
\begin{equation}
\hat{r}_t
=
f_{\Theta}
\left(
x_t,
\mathcal{H}^{s_t}_{<\tau_t},
\mathcal{R}_t,
\Gamma_t
\right),
\end{equation}
where $\Gamma_t$ remains available during inference, while the predictive
influence of the retrieved evidence is determined adaptively by
$f_{\Theta}(\cdot)$.
%

\section{Proposed Method}
\begin{figure*}[!ht]
    \centering
    \resizebox{0.98\textwidth}{7.5 cm}{%
        \includegraphics{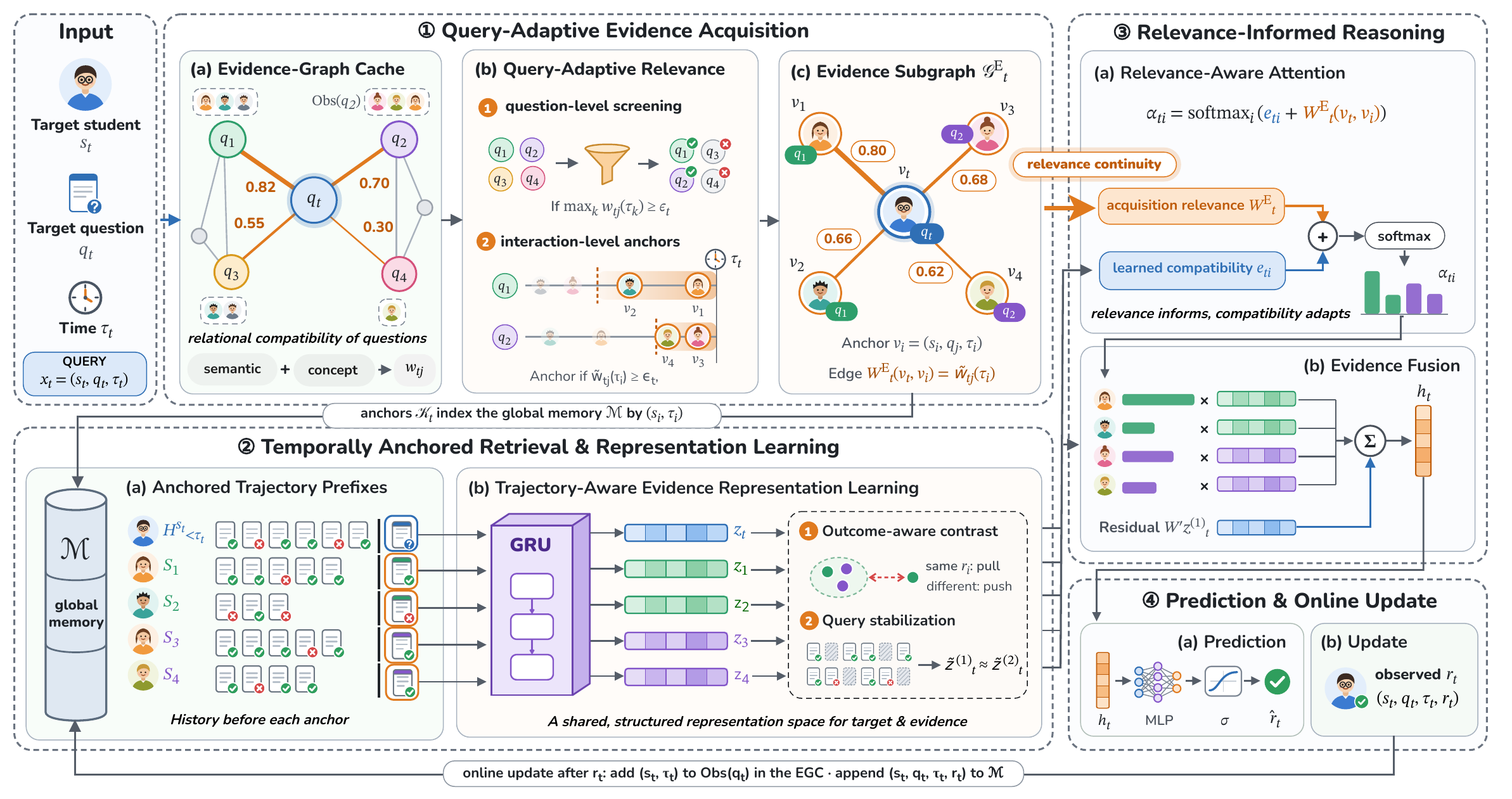}%
    }
    \caption{Overall framework of CaRe-KT, which realizes relevance continuity from evidence acquisition to knowledge-state inference. \textcircled{1} The Evidence-Graph Cache refines persistent question-level compatibility through query-adaptive candidate screening and temporally valid interaction-anchor selection, yielding a query-specific evidence subgraph whose edge weights preserve acquisition relevance.
    \textcircled{2} Selected anchors guide temporally grounded trajectory retrieval, while a shared encoder and complementary regularizers establish a behaviorally structured representation space for the retrieved trajectories and target history.
    \textcircled{3} The preserved relevance enters attention alongside learned query-evidence compatibility (orange arrow), informing the influence of each trajectory without determining it. \textcircled{4} The fused representation yields $\hat{r}_t$, and the observed response updates the cache and memory. Numerical values are illustrative.}
    \label{fig:placeholder}
\end{figure*}

In this section, we introduce CaRe-KT, a Continuity-aware Retrieval-augmented Reasoning framework for
KT that maintains relevance continuity from evidence acquisition to downstream reasoning while allowing the predictive influence of retrieved evidence to remain context-adaptive. At its core, the Evidence-Graph Cache (EGC) is a dynamically maintained weighted question graph that encodes persistent semantic-conceptual relationships among observed questions while maintaining per-question historical occurrence buffers. Fig. 2 illustrates its query-level workflow. For each incoming query, CaRe-KT activates a compatibility-based neighborhood around the target question in the EGC and refines it according to the temporal validity of associated historical occurrences, yielding a temporally grounded, interaction-level evidence subgraph. Each subgraph node serves as a historical interaction anchor, indexing the global interaction memory to retrieve the learner trajectory preceding that interaction. The retrieved trajectories, together with the current query trajectory, are then encoded in a shared representation space, where an outcome- and stage-aware contrastive regularization shapes its geometry and query-consistency regularization stabilizes the query representation under history perturbations. Rather than discarding the relevance information used in evidence selection, CaRe-KT carries the resulting subgraph weights into attention, where they interact with learned query–evidence compatibility to determine the relative influence of historical trajectory representations. The aggregated evidence supports correctness prediction, followed by causal cache and memory updates upon observing the actual response.
\subsection{Dynamic Relevance Modeling}
CaRe-KT models query-dependent behavioral relevance through two complementary factors: \textit{relational compatibility}, which captures persistent semantic and conceptual relations between questions, and \textit{temporal validity}, which reflects the query-specific usefulness of historical evidence over time. These factors are organized through the \textbf{Evidence-Graph Cache (EGC)}, a dynamically maintained weighted question graph that supports query-adaptive evidence acquisition.

At time $\tau_t$, the historical EGC is represented as a dynamic graph $G^t=(\mathcal{V}^t,\mathcal{E}^t,\mathcal{W}^t)$, where $\mathcal{V}^t\subseteq\mathcal{Q}_{<\tau_t}$ contains the active historically observed questions, $\mathcal{E}^t\subseteq\mathcal{V}^t\times\mathcal{V}^t$ encodes undirected question relations, and $\mathcal{W}^t=[w_{ij}]$ stores their relational compatibilities. Each node $q_i\in\mathcal{V}^t$ is associated with a semantic representation $\mathbf{e}^{sem}_i$, a concept set $\mathcal{C}_i$, and an interaction buffer \(\mathrm{Obs}(q_i)=\{(s_k,\tau_k)\mid q_k=q_i,\ \tau_k<\tau_t\}\) , containing only its historical occurrences. 

The relational compatibility between $q_i$ and $q_j$ is defined by combining semantic similarity and conceptual overlap
\begin{equation}
\label{eq:relational_compatibility}
w_{ij}
=
\alpha_{ij}\,
\mathrm{Cos}^{+}(\mathbf{e}^{sem}_i,\mathbf{e}^{sem}_j)
+
(1-\alpha_{ij})\,
\mathrm{Jac}(\mathcal{C}_i,\mathcal{C}_j),
\end{equation}
where $\mathrm{Cos}^{+}(\cdot,\cdot)$ and $\mathrm{Jac}(\cdot,\cdot)$ denote cosine similarity normalized to $[0,1]$ and Jaccard similarity, respectively. The mixing coefficient $\alpha_{ij}$ is determined by the degree of conceptual agreement:
\begin{equation}
\label{eq:adaptive_blending}
\alpha_{ij}
=
\sigma\!~(
\beta_0-\beta_1
\frac{2|\mathcal{C}_i\cap\mathcal{C}_j|}
{|\mathcal{C}_i|+|\mathcal{C}_j|}),
\end{equation}
where $\sigma(\cdot)$ is the sigmoid function, and $\beta_0$ and $\beta_1$ are learnable parameters, with $\beta_1\ge 0$. Consequently, stronger concept agreement increases the contribution of conceptual similarity, whereas weaker agreement places greater emphasis on semantic similarity.

For an incoming query $x_t=(s_t,q_t,\tau_t)$, CaRe-KT activates a query-centered EGC view. If $q_t\notin\mathcal{V}^t$, it is added to the node set, $\mathcal{V}^t\leftarrow\mathcal{V}^t\cup\{q_t\}$, with semantic representation $\mathbf{e}^{sem}_t$, concept set $\mathcal{C}_t$, and an empty interaction buffer $\mathrm{Obs}(q_t)=\emptyset$. Its relational compatibilities $w_{ti}$ with the existing historical nodes are then computed using Eq.~\eqref{eq:relational_compatibility}; otherwise, its existing node, historical observations, and relational compatibilities are retained. 
While $w_{ti}$ represents persistent question-level compatibility, individual occurrences of $q_i$ may differ in their temporal relevance to the current query. For each $(s_k,\tau_k)\in\mathrm{Obs}(q_i)$, CaRe-KT therefore defines
%
\begin{equation}
\label{eq:dynamic_relevance}
\tilde{w}_{ti}(\tau_k)
=
w_{ti}\exp\!\left[-\lambda(\tau_t-\tau_k)\right],
\end{equation}
where $\lambda>0$ controls temporal decay. This refinement preserves the underlying question relation while differentiating historical interactions according to their temporal proximity to the prediction context. All eligible occurrences satisfy $\tau_k<\tau_t$, ensuring that relevance estimation uses only historical information.
\subsection{Query-Adaptive Evidence Subgraph Construction}
Given the dynamic relevance model, CaRe-KT constructs a query-specific evidence subgraph through hierarchical refinement: question-level screening identifies a relevant region of the EGC, while interaction-level filtering resolves candidate questions into temporally valid historical anchors.

\vspace{0.1 cm}
\noindent\textbf{1. Question-Level Candidate Identification.} For the incoming query $x_t=(s_t,q_t,\tau_t)$, CaRe-KT first localizes the region of the historical question space that may contain relevant evidence.
For each $q_j\in\mathcal{V}^t$ with nonempty $\mathrm{Obs}(q_j)$, let
\begin{equation}
\label{eq:latest_occurrence}
\tau_j^{\max}
=
\max\{\tau_k\mid(s_k,\tau_k)\in\mathrm{Obs}(q_j)\}
\end{equation}
denote its most recent historical occurrence. Since Eq.~\eqref{eq:dynamic_relevance} decreases monotonically with temporal distance, $\tilde{w}_{tj}(\tau_j^{\max})$ gives the maximum occurrence-level relevance of $q_j$ to the current query. CaRe-KT defines a query-adaptive relevance boundary
\begin{equation}
\label{adaptivebound}
\epsilon_t
=
\eta
\max_{q_j\in\mathcal{V}^t}
\tilde{w}_{tj}(\tau_j^{\max}),
\end{equation}
where $\eta\in(0,1)$ controls relative selectivity. The candidate question set is then given by
\begin{equation}
\label{eq:candidate_question_set}
\mathcal{N}_{\epsilon}(q_t)
=
\left\{
q_j\in\mathcal{V}^t
\mid
\tilde{w}_{tj}(\tau_j^{\max})\geq\epsilon_t
\right\}.
\end{equation}
Unlike fixed-cardinality retrieval, this criterion adapts the candidate region to the query-specific relevance distribution.

\vspace{0.1 cm}
\noindent\textbf{2. Interaction-Level Anchor Identification.} Each candidate question may have multiple historical occurrences with different temporal relevance. For each $q_j\in\mathcal{N}_{\epsilon}(q_t)$,  CaRe-KT retains an occurrence $(s_k,\tau_k)\in\mathrm{Obs}(q_j)$ as an evidence anchor if
\begin{equation}
\label{eq:interaction_anchor_condition}
\tilde{w}_{tj}(\tau_k)\geq\epsilon_t.
\end{equation}
The resulting anchor set is
\begin{equation}
\resizebox{\columnwidth}{!}{$
\mathcal{K}_t
=
\left\{
(s_k,q_j,\tau_k)
\;\middle|\;
q_j\in\mathcal{N}_{\epsilon}(q_t),
\;
(s_k,\tau_k)\in\mathrm{Obs}(q_j),
\;
\tilde{w}_{tj}(\tau_k)\geq\epsilon_t
\right\}
$}
\end{equation}
These anchors identify temporally relevant historical interactions rather than merely related questions.

\noindent\textbf{3. Evidence Subgraph Construction.}
The selected anchors and incoming query define a query-centered weighted evidence subgraph
\begin{equation}
\label{eq:evidence_subgraph}
\mathcal{G}_t^E=
(\mathcal{V}_t^E,\mathcal{E}_t^E,\mathcal{W}_t^E),
\end{equation}
where $\mathcal{V}_t^E=\{v_t\}\cup\{v_i=(s_i,q_j,\tau_i)\mid(s_i,q_j,\tau_i)\in\mathcal{K}_t\}$, with $v_t=x_t =(s_t,q_t,\tau_t)$. Each anchor is connected to the query node through an edge $(v_t,v_i)\in\mathcal{E}_t^E$, weighted by
\begin{equation}
\label{eq:evidence_edge_weight}
\mathcal{W}_t^E(v_t,v_i)
=
\tilde{w}_{tj}(\tau_i).
\end{equation}
The resulting subgraph $\mathcal{G}^{E}_t$ retains interaction-level acquisition relevance alongside the selected historical anchors, providing the shared evidence structure for subsequent retrieval and relevance-informed reasoning.
%
\subsection{Temporally Anchored Behavioral Evidence Retrieval}
Given the query-specific evidence subgraph $\mathcal{G}_t^{E}$, each historical anchor node $v_i=(s_i,q_i,\tau_i)\in\mathcal{K}_t$ identifies the student history to be accessed and the temporal boundary at which it should be retrieved. CaRe-KT then extracts the prefix of student $s_i$'s interaction history observed before $\tau_i$ from the global interaction memory $\mathcal{M}_{<\tau_t}$:
\begin{equation}
\label{eq:retrieval}
\mathcal{S}_i
=
\left\{
(q_{\ell},\tau_{\ell},r_{\ell})
\in
\mathcal{H}^{s_i}_{<\tau_t}
\mid
\tau_{\ell}<\tau_i
\right\}.
\end{equation}
Retrieval at $\tau_i$ preserves the behavioral context available prior to the anchored interaction, preventing subsequent interactions of the same student from influencing its representation. Distinct anchors from the same student may therefore yield different historical prefixes, capturing different stages of learning.

The retrieved behavioral evidence is collected as
\begin{equation}
\label{eq:retrieved_evidence}
\mathcal{R}_t
=
\left\{
(\mathcal{S}_i,\tau_i,r_i)
\mid
v_i=(s_i,q_i,\tau_i)\in\mathcal{K}_t
\right\},
\end{equation}
where $r_i$ is the historically observed response at anchor $v_i$. The query node $v_t=(s_t,q_t,\tau_t)$ is represented using the target learner's causal history $\mathcal{H}^{s_t}_{<\tau_t}$, without accessing the unknown response $r_t$.
\subsection{Trajectory-Aware Evidence Representation Learning}

Each retrieved anchor $v_i=(s_i,q_i,\tau_i)$ associates a preceding learner trajectory $\mathcal{S}_i$ with an observed response $r_i$ that characterizes performance at a particular stage of learning. CaRe-KT encodes these trajectories and the target learner's history into a shared behavioral representation space, where complementary regularizers promote outcome-discriminative structure and stabilize learner-history representations for subsequent knowledge-state inference.

Each interaction $(q_{\ell},\tau_{\ell},r_{\ell})\in\mathcal{S}_i$ is represented by combining question content with its observed response:
\begin{equation}
\label{eq:interaction_representation}
\tilde{\mathbf{h}}_{\ell}
=
\mathbf{v}_{q_{\ell}}
\oplus
\mathbf{e}^{res}_{\ell},
\end{equation}
where $\oplus$ denotes concatenation and the question representation integrates semantic and conceptual information:
\begin{equation}
\label{eq:question_representation}
\mathbf{v}_{q_{\ell}}
=
\mathbf{W}_{s}\mathbf{e}^{sem}_{\ell}
+
\mathbf{W}_{c}\mathbf{e}^{con}_{\ell}.
\end{equation}
Here, $\mathbf{e}^{con}_{\ell}$ is the multi-hot concept representation of $q_{\ell}$, $\mathbf{W}_{s}$ and $\mathbf{W}_{c}$ are learnable projections, and $\mathbf{e}^{res}_{\ell}=\mathrm{OneHot}(r_{\ell})\mathbf{B}$ is the observed-response representation, with the embedding matrix $\mathbf{B}\in\mathbb{R}^{2\times d_r}$. 

The resulting interaction representations are processed chronologically by a shared GRU encoder $f_{\theta}$. For a trajectory $\mathcal{S}_i$ of length $L_i$, its behavioral representation is
\begin{equation}
\label{eq:trajectory_encoding}
\mathbf{z}_i
=
f_{\theta}
\left(
\tilde{\mathbf{h}}_{\ell_1},
\ldots,
\tilde{\mathbf{h}}_{\ell_{L_i}}
\right),
\end{equation}
where $\ell_1,\ldots,\ell_{L_i}$ follow the temporal order of $\mathcal{S}_i$. The same encoder produces $\mathbf{z}_t=f_{\theta}(\mathcal{H}^{s_t}_{<\tau_t})$ from the target learner's causal history. To condition this representation on the target question, CaRe-KT constructs
\begin{equation}
\label{eq:target_query_representation}
\mathbf{z}_t^q
=
\mathrm{LayerNorm}
\left(
\mathbf{W}_{Q}
[\mathbf{z}_t\Vert\mathbf{v}_{q_t}]
\right),
\end{equation}
where $\mathbf{v}_{q_t}$ follows Eq.~\eqref{eq:question_representation}, and $\mathbf{W}_{Q}$ projects the concatenated learner-history and target-question representations into the reasoning space. Thus, $\mathbf{z}_t$ encodes historical learner behavior, whereas $\mathbf{z}_t^q$ provides the question-conditioned representation used for downstream evidence reasoning. Two complementary regularizers further structure the behavioral space and stabilize the learner-history encoding.

\vspace{0.1 cm}
\noindent\textbf{a. Trajectory-Aware Contrastive Regularization.}
Historical anchor responses provide supervision for distinguishing behavioral trajectories.
CaRe-KT leverages this anchor-level supervision through a trajectory-aware supervised contrastive regularizer that shapes the geometry of the retrieved trajectory representation space by increasing representational compatibility among trajectories sharing the same anchor outcome, while temporal proximity modulates their pairwise influence. For each anchor $v_i$, the positive set is
\begin{equation}
\label{eq:positive_anchor_set}
P(v_i)
=
\left\{
v_j\in\mathcal{K}_t
\mid
v_j\neq v_i,\ r_j=r_i
\right\}.
\end{equation}
Let $\mathcal{I}_t^+=\{v_i\in\mathcal{K}_t\mid P(v_i)\neq\emptyset\}$ denote anchors with at least one positive counterpart. 
The supervised contrastive objective is then defined as
\begin{align} \label{SCLloss} 
\mathcal{L}_{\mathrm{SCL}} = 
\sum_{v_i\in\mathcal{I}_t^{+}} \frac{-1}{|P(v_i)|} \sum_{v_j\in P(v_i)} \log \frac{ \exp\!\left(h(\mathbf{z}_i,\mathbf{z}_j)/\zeta\right) }{ \sum_{\substack{v_k\neq v_i}} \exp\!\left(h(\mathbf{z}_i,\mathbf{z}_k)/\zeta\right) }, \end{align}
where $\zeta>0$ is the temperature parameter. 
The similarity function $h(\mathbf{z}_i,\mathbf{z}_j)$ combines representational similarity with temporal proximity:
\begin{equation}
\label{eq:trajectory_similarity}
h(\mathbf{z}_i,\mathbf{z}_j)
=
\mathrm{Cos}^{+}(\mathbf{z}_i,\mathbf{z}_j)
\exp\!\left(-\beta|\tau_i-\tau_j|\right),
\end{equation}
where $\beta\geq0$ controls temporal sensitivity. This formulation promotes outcome-discriminative behavioral representations while reducing the similarity contribution of temporally distant anchors.

\vspace{0.1 cm}
\noindent\textbf{b. Query Representation Stabilization.}
Although contrastive regularization structures the retrieved behavioral representations, perturbations of the target learner's history may alter its encoded state and affect subsequent question-conditioned reasoning. CaRe-KT therefore introduces a consistency objective over the target learner's causal history. 
Given $\mathcal{H}^{s_t}_{<\tau_t}$, two order-preserving views $\tilde{\mathcal{H}}_t^{(1)}$ and $\tilde{\mathcal{H}}_t^{(2)}$ are generated through random interaction masking~\cite{xie2022contrastive} and encoded by the shared $f_{\theta}$ to obtain $\tilde{\mathbf{z}}_t^{(1)}$ and $\tilde{\mathbf{z}}_t^{(2)}$. Their consistency is encouraged through a symmetric objective inspired by SimSiam~\cite{chen2021exploring}:
\begin{equation}
\label{eq:query_stabilization}
\begin{aligned}
\mathcal{L}_{\mathrm{self}}
={}&
\frac{1}{2}\mathcal{D}
\left(
\tilde{\mathbf{z}}_t^{(1)},
\tilde{\mathbf{z}}_t^{(2)}
\right)
+
\frac{1}{2}\mathcal{D}
\left(
\tilde{\mathbf{z}}_t^{(2)},
\tilde{\mathbf{z}}_t^{(1)}
\right),
\end{aligned}
\end{equation}
where $\mathcal{D}(\cdot,\cdot)$ denotes negative cosine similarity with stop-gradient applied to the target representation in each directional term. By enforcing local consistency across perturbed views of the same causal history, this objective stabilizes the position of the central query representation within the contrastively structured behavioral space.

The combined representation regularization objective is
\begin{equation}
\label{eq:representation_regularization}
\mathcal{L}_{\mathrm{reg}}
=
\lambda_{\mathrm{SCL}}\mathcal{L}_{\mathrm{SCL}}
+
\lambda_{\mathrm{self}}\mathcal{L}_{\mathrm{self}},
\end{equation}
where $\lambda_{\mathrm{SCL}}$ and $\lambda_{\mathrm{self}}$ control the respective regularization strengths.
\subsection{Query-Conditioned Evidence Reasoning}
The query-specific evidence subgraph $\mathcal{G}_t^{E}$ preserves both the selected behavioral evidence and its acquisition-derived relevance. CaRe-KT reasons over this structure by combining persistent evidence relevance with learned query-evidence compatibility, allowing retrieved trajectories to contribute according to the current learner state and target question.

\vspace{0.1 cm}
\noindent\textbf{a. Query-Conditioned Evidence Fusion.}
Let $\mathcal{N}_t^{E}$ denote the historical anchor nodes connected to the query node $v_t$. CaRe-KT integrates their behavioral representations through query-conditioned attention, yielding
\begin{equation}
\label{eq:evidence_fusion}
\mathbf{h}_t
=
\sigma\!~(
\mathbf{W}'\mathbf{z}_t^q
+
\sum_{v_i\in\mathcal{N}_t^{E}}
\alpha_{ti}\mathbf{W}\mathbf{z}_i ),
\end{equation}
where the residual pathway retains the question-conditioned query representation $\mathbf{z}_t^q$, while $\alpha_{ti}$ controls the relative influence of each retrieved trajectory. To maintain relevance continuity, the attention weights incorporate the acquisition-derived edge relevance $\mathcal{W}_t^{E}(v_t,v_i)$ alongside learned query-evidence compatibility:
\begin{equation}
\label{eq:evidence_attention}
\alpha_{ti}
=
\frac{
\exp\!\left(
e_{ti}
+
\mathcal{W}_t^{E}(v_t,v_i)
\right)
}{
\displaystyle
\sum_{v_j\in\mathcal{N}_t^{E}}
\exp\!\left(
e_{tj}
+
\mathcal{W}_t^{E}(v_t,v_j)
\right)
},
\end{equation}
where $e_{ti}=\mathrm{LeakyReLU}\!\left(\mathbf{a}^{\top}[\mathbf{W}\mathbf{z}_t^q\Vert\mathbf{W}\mathbf{z}_i]\right)$ denotes learned compatibility between the current query and historical evidence. Consequently, acquisition relevance remains available as a structural prior during reasoning, while learned compatibility adapts the relative contribution of retrieved trajectories to the current prediction context.

\vspace{0.1 cm}
\noindent\textbf{b. Prediction.}
The evidence-conditioned representation $\mathbf{h}_t$ is mapped to the probability of a correct response to question $q_t$ by student $s_t$:
\begin{equation}
\label{eq:prediction}
\hat{r}_t
=
\sigma\!\left(
\mathrm{MLP}(\mathbf{h}_t)
\right),
\end{equation}
where $\hat{r}_t\in[0,1]$. Prediction is then supervised through the binary cross-entropy objective
\begin{equation}
\label{eq:prediction_loss}
\mathcal{L}_{\mathrm{pred}}
=
-\sum_t
\left[
r_t\log\hat{r}_t
+
(1-r_t)\log(1-\hat{r}_t)
\right].
\end{equation}
%
\subsection{Dynamic Cache and Memory Maintenance}
After observing $r_t$, CaRe-KT updates $\mathrm{Obs}(q_t)\leftarrow\mathrm{Obs}(q_t)\cup\{(s_t,\tau_t)\}$ and appends $(s_t,q_t,\tau_t,r_t)$ to the global interaction memory $\mathcal{M}$, making the interaction available for subsequent queries. To bound the active evidence space, the EGC is periodically pruned by retaining only occurrences within a temporal horizon $W$, i.e., $\mathrm{Obs}(q_i)\leftarrow\{(s_k,\tau_k)\in\mathrm{Obs}(q_i)\mid\tau_k\geq\tau_t-W\}$. Nodes with empty observation buffers and edges with negligible relational compatibility are also removed. While temporal decay continuously attenuates the relevance of older occurrences, pruning removes outdated evidence from the active EGC. The global interaction memory remains unpruned, preserving complete learner histories for retrieval through EGC-selected anchors.
%
%
\subsection{Model Learning and Optimization}

Let $\Theta$ denote the trainable parameters of CaRe-KT. Model learning jointly optimizes response prediction and behavioral representation regularization:
\begin{equation}
\label{eq:total_loss}
\Theta^{*}
=
\arg\min_{\Theta}
\left(
\mathcal{L}_{\mathrm{pred}}
+
\mathcal{L}_{\mathrm{reg}}
\right).
\end{equation}
Gradients are propagated through all differentiable components involved in the prediction process.
\subsection{Computational Complexity}
Let $N_t=|\mathcal{V}^{t}|$, $M_t=|\mathcal{N}_{\epsilon}(q_t)|$, and $K_t=|\mathcal{K}_t|$ denote the numbers of active EGC questions, candidate questions, and selected anchors, respectively, with $b_j=|\mathrm{Obs}(q_j)|$. Given stored relational compatibilities and temporally ordered observation buffers, question-level screening costs $\mathcal{O}(N_t)$. The monotonic temporal decay in Eq.~\eqref{eq:dynamic_relevance} enables binary-search-based anchor identification, yielding an evidence-subgraph construction cost of $\mathcal{O}(\sum_{j=1}^{M_t}\log b_j+K_t)$.
Let $d$ denote the representation dimension, $L_t$ the target history length, and $S_t=\sum_{v_i\in\mathcal{K}_t}|\mathcal{S}_i|$ the total retrieved trajectory length. Sequential encoding and evidence fusion require $\mathcal{O}((S_t+L_t+K_t)d^2)$, while training additionally incurs $\mathcal{O}(K_t^2d)$ for contrastive regularization. Hence, per-query inference requires $\mathcal{O}(N_t+\sum_{j=1}^{M_t}\log b_j+(S_t+L_t+K_t)d^2)$, with an additional $\mathcal{O}(K_t^2d)$ during training. Previously encoded historical trajectories can be reused at inference when their representations are cached under fixed encoder parameters. For unseen questions, computing EGC compatibilities adds $\mathcal{O}(N_t d_q)$, where $d_q$ denotes the per-pair compatibility evaluation cost.

\begin{algorithm}[!t]
%
\caption{CaRe-KT Query-Level Inference} \label{alg:care_kt}
\centering
\scalebox{0.95}{%
\begin{minipage}{\columnwidth}
\small
\begin{algorithmic}[1]
\Require Chronological queries $x_t=(s_t,q_t,\tau_t)$;
historical memory $\mathcal{M}_{<\tau_t}$; EGC $G^{t}$; trained parameters $\Theta$
\Ensure Predicted correctness probabilities $\{\hat r_t\}$
\For{each query $x_t$ in chronological order}
     \vspace{0.1 cm}
    \Statex \hspace{\algorithmicindent}\(\triangleright\)~\textbf{Phase I: Query-Adaptive Evidence acquisition}
    \vspace{0.05 cm}
    \If{$q_t\notin\mathcal{V}$}
        \State Add $q_t$ to $G^t$; compute its compatibilities \Comment{Eqs.~(4)--(5)}
    \EndIf
    \State Compute adaptive relevance boundary $\epsilon_t$ \Comment{Eq.~(\ref{adaptivebound})}
    \State Select candidate questions $\mathcal{N}_\epsilon(q_t)$ \Comment{Eq.~(\ref{eq:candidate_question_set})}
    \State Identify temporally valid anchors $\mathcal{K}_t$ \Comment{Eqs.~(10)--(11)}
    \State Build subgraph $\mathcal{G}^E_t$ with weights $\mathbf{W}^E_t$ \Comment{Eqs.~(12)--(13)}
    \vspace{0.1 cm}
    \Statex \hspace{\algorithmicindent}\(\triangleright\)~\textbf{Phase II: Behavior-Aware Trajectory Encoding}
    \vspace{0.05 cm}
    \For{each anchor $v_i\in\mathcal{K}_t$}
        \State Retrieve preceding trajectory $\mathcal{S}_i$ from $\mathcal{M}$ \Comment{Eq.~(14)}
        \State Encode $\mathbf{z}_i\gets f_\theta(\mathcal{S}_i)$ \Comment{Eqs.~(16)--(18)}
    \EndFor
    \State Encode target history $\mathbf{z}_t\gets f_\theta(\mathcal{H}^{s_t}_{<\tau_t})$
    \State $\mathbf{z}^q_t\gets\mathrm{LayerNorm}\big(\mathbf{W}_Q[\mathbf{z}_t\,\|\,\mathbf{v}_{q_t}]\big)$ \Comment{Eqs.~(\ref{eq:question_representation}),~(\ref{eq:target_query_representation})}
    \vspace{0.1 cm}
    \Statex \hspace{\algorithmicindent}\(\triangleright\)~\textbf{Phase III: Relevance-Informed Reasoning}
    \vspace{0.05 cm}
    \State Compute learned compatibility $e_{ti},\ \forall v_i\in\mathcal{K}_t$ \Comment{Eq.~(26)}
    \State $\alpha_{ti}\gets\mathrm{softmax}_i\big(e_{ti}+\mathbf{W}^E_t(v_t,v_i)\big)$ \Comment{relevance continuity}
    \State Aggregate $\mathbf{h}_t$ from $\mathbf{z}_t^q$ and $\{\alpha_{ti}\mathbf{z}_i\}$ \Comment{Eq.~(\ref{eq:evidence_fusion})}
    \State Predict $\hat{r}_t$ from $\mathbf{h}_t$ \Comment{Eq.~(27)}
    \vspace{0.05 cm}
    \Statex \hspace{\algorithmicindent}\(\triangleright\)~\textbf{Phase IV: Causal update}
    \vspace{0.05 cm}
    \State Observe $r_t$; append $(s_t,\tau_t)$ to $\mathrm{Obs}(q_t)$ and $(x_t,r_t)$ to $\mathcal{M}$
\EndFor
\end{algorithmic}
\end{minipage}%
}
\end{algorithm}
\section{Experiments}
In this section, we empirically evaluate CaRe-KT through five research questions. RQ1: How does CaRe-KT perform compared with state-of-the-art KT models across diverse datasets? RQ2: How robust is CaRe-KT under sparse and cold-start conditions? RQ3: How does relevance continuity, through combining acquisition-derived relevance with learned query–evidence compatibility, affect CaRe-KT’s predictive performance? RQ4: How effectively does the EGC construct query-adaptive and temporally valid behavioral evidence for prediction? RQ5: How does behaviorally structured representation learning affect CaRe-KT’s predictive performance? We further conduct sensitivity analyses to examine the robustness of CaRe-KT to its key hyperparameters.
\subsection{Experimental Setup}

\subsubsection{Datasets}
We evaluate CaRe-KT on five public benchmarks (Table~\ref{tab:datasets}) chosen to reflect a range of conditions under which adaptive learning systems operate.
DBE-KT22\footnote{\url{https://doi.org/10.26193/6DZWOH}} was collected from a university database course and Slepemapy.cz\footnote{\url{https://www.fi.muni.cz/adaptivelearning/data/slepemapy/}} from an adaptive system for practising geography facts, while ASSISTment09\footnote{Obtained through EduData, \url{https://github.com/bigdata-ustc/EduData}.\label{fn:edudata}}, XES3G5M\footnote{\url{https://github.com/ai4ed/XES3G5M}}, and Junyi\textsuperscript{\ref{fn:edudata}} were collected from online mathematics platforms for school-age learners. Beyond domain and scale, they differ in two properties central to retrieval-augmented KT. Cross-learner evidence per question is sparsest on ASSISTment09 and densest on Junyi, and question text is available only on DBE-KT22 and XES3G5M, with descriptive metadata elsewhere.
\begin{table}[!t]
    \caption{Statistical Properties of the Evaluation Datasets.} \label{tab:datasets}
    \centering
    \scalebox{0.95}{\begin{tabular}{lrrrr}
        \toprule
        Dataset & Students & Questions & Concepts & Interactions \\
        \midrule
        DBE-KT22     & 1.3k   & 212   & 93    & 162.0k     \\
        ASSISTment09 & 4.2k   & 26.7k & 110   & 401.8k     \\
        XES3G5M      & 18.1k  & 7.7k  & 865   & 5,549.6k   \\
        Slepemapy.cz & 91.3k  & 4.5k  & 1,458 & 10,087.3k  \\
        Junyi        & 247.6k & 0.7k  & 39    & 25,926.0k \\
        \bottomrule
    \end{tabular}}
\end{table}
\subsubsection{Baselines}
We compare CaRe-KT with thirteen representative KT models in three groups:
\begin{itemize}
    \setlength{\itemsep}{0pt}
    \item \textbf{Sequential state-based models.} DKT~\cite{piech2015deep}, together with IEKT~\cite{long2021tracing}, DIMKT~\cite{shen2022assessing}, and QIKT~\cite{chen2023improving}, which extend recurrent state modeling with individual cognition, question difficulty, and question-centric representations, respectively.
    \item \textbf{Attention- and graph-based models.} AKT~\cite{ghosh2020context}, simpleKT~\cite{liu2023simplekt}, stableKT~\cite{li2024enhancing}, and UKT~\cite{cheng2025uncertainty}, which weight past interactions through attention, while DyGKT~\cite{cheng2024dygkt} models learner--question interactions as a dynamic graph.
    \item \textbf{Collaborative and retrieval-augmented models.} CoKT~\cite{long2022improving}, CoSKT~\cite{zhang2024coskt}, RACL~\cite{berahmand2026retrieval}, and FINER~\cite{liu2025advancing}, which draw on evidence from other learners and are the most direct comparison for CaRe-KT.
\end{itemize}
%
\subsubsection{Evaluation Metrics}
We evaluate predictive performance with three metrics widely adopted in KT~\cite{cheng2024dygkt, li2025hyperbolic}: the area under the ROC curve (AUC), average precision (AP), and root mean squared error (RMSE). AUC and AP assess how well correct responses are ranked above incorrect ones, whereas RMSE measures the error of the predicted probabilities.

\subsubsection{Implementation Details}
All models are evaluated under a global chronological data split~\cite{cheng2024dygkt}: interactions are sorted by timestamp and split 80\%/10\%/10\% into training, validation, and test sets, so that testing involves strictly future interactions. For the question-level cold-start scenario, questions covering 10\% of the interactions are withheld from training and validation and used only for testing (averaged over five random subsets on DBE-KT22, which has only 212 questions). Every model is scored on the same test interactions under a common online protocol, in which parameters are frozen after training and each scored interaction becomes available to all models thereafter; the collaborative baselines retrieve from the same global memory $\mathcal{M}_{<\tau_t}$ as CaRe-KT. Baselines use official implementations, or pyKT~\cite{liu2022pykt} for the sequential and attention-based models, with recommended hyperparameters and only the minimal corrections this protocol requires (e.g., removing a label leak in FINER). Question content is the question text on DBE-KT22 and XES3G5M. The other datasets provide no question text, so a short description is composed from metadata, namely the question identifier and skill names on ASSISTment09, the place name, place type, map, and question type on Slepemapy.cz, and the exercise name, topic, and area on Junyi. All content is embedded with a frozen all-MiniLM-L6-v2 encoder~\cite{reimers2019sentence}. We use dimension 128, histories of up to 200 interactions, at most 20 anchors, $\eta=0.75$, $\lambda=\ln 2/80$ and $W=50$ interactions, and $\lambda_{\mathrm{SCL}}=\lambda_{\mathrm{self}}=1$, and train with Adam (learning rate $10^{-3}$, batch size 128, dropout 0.1) for up to 100 epochs with early stopping (patience 10). This configuration was selected on DBE-KT22 and ASSISTment09 validation data and used for all datasets. For every model, the checkpoint with the lowest validation loss is tested, with a fixed random seed, on NVIDIA A100 and H100 GPUs with PyTorch 2.11.

\subsection{Results and Analysis}
\subsubsection{RQ1. Overall Predictive Performance}
\begin{table*}[!ht]
    \caption{Overall performance on five datasets. The best result in each column is shaded dark blue and in bold, while the second best is shaded light blue and underlined. Arrows indicate the preferred direction ($\uparrow$ higher is better, $\downarrow$ lower is better)}
    \label{OveralPerformance}
    \centering
    \scalebox{0.85}{\begin{tabular}{lccccccccccccccccccc}
    \hline
    \multirow{2}{*}{\textbf{Model}} & \multicolumn{3}{c}{\textbf{DBE-KT22}} & & \multicolumn{3}{c}{\textbf{ASSISTment09}} && \multicolumn{3}{c}{\textbf{Slepemapy.cz}} && \multicolumn{3}{c}{\textbf{Junyi}} && \multicolumn{3}{c}{\textbf{XES3G5M}} \\
    \cline{2-4} \cline{6-8} \cline{10-12}\cline{14-16} \cline{18-20}
        & $\uparrow$AUC & $\uparrow$AP & $\downarrow$RMSE &&  $\uparrow$AUC & $\uparrow$AP & $\downarrow$RMSE && $\uparrow$AUC & $\uparrow$AP & $\downarrow$RMSE && $\uparrow$AUC & $\uparrow$AP & $\downarrow$RMSE && $\uparrow$AUC & $\uparrow$AP & $\downarrow$RMSE \\
    \hline
    DKT
        & 76.55 & 89.72 & 0.4031 &
        & 69.45 & 84.08 & 0.444 &
        & 64.62 & 83.94 & 0.3861 &
        & 69.01 & 90.70 & 0.3612 &
        & 65.14 & 88.85 & 0.4007 \\

    AKT
        & 76.53 & 89.75 & 0.4080 &
        & 77.43 & 88.31 & 0.423 &
        & 67.39 & 85.59 & 0.3802 &
        & 73.10 & 91.72 & 0.3549 &
        & 69.12 & 89.42 & 0.3871 \\

    IEKT
        & 79.11 & 91.14 & \best{0.3894} &
        & 76.97 & \second{88.74} & 0.477 &
        & 69.87 & 87.00 & 0.3795 &
        & 77.09 & 92.76 & 0.3514 &
        & 72.49 & 91.09 & 0.3655 \\

    DIMKT
        & 75.56 & 89.62 & 0.4174 &
        & 75.53 & 88.69 & 0.452 &
        & 68.09 & 84.03 & 0.3793 &
        & 72.51 & 91.02 & 0.3556 &
        & 72.75 & 90.70 & 0.3624 \\

    QIKT
        & 79.59 & 91.14 & 0.3912 &
        & 71.98 & 85.54 & 0.434 &
        & 69.79 & 86.97 & 0.378 &
        & 75.38 & 92.09 & 0.3575 &
        & 66.65 & 88.88 & 0.4026 \\

    simpleKT
        & 75.08 & 89.26 & 0.4189 &
        & 65.52 & 78.89 & 0.472 &
        & 68.62 & 86.51 & 0.3801 &
        & 76.98 & 92.84 & 0.3544 &
        & 69.78 & 89.86 & 0.3752 \\

    DyGKT
        & 79.44 & 91.05 & 0.3909 &
        & 70.35 & 85.57 & 0.462 &
        & 72.49 & 88.39 & 0.3763 &
        & 80.35 & 94.01 & 0.3411 &
        & \second{75.39} & \second{92.40} & \second{0.3606} \\

    stableKT
        & 75.48 & 89.55 & 0.4174 &
        & \second{81.45} & 87.06 & \best{0.3786} &
        & 77.53 & 92.08 & 0.377 &
        & 80.56 & \second{94.36} & 0.3394 &
        & 70.43 & 90.27 & 0.3783 \\

    UKT
        & 75.82 & 89.46 & 0.4098 &
        & 74.78 & 86.84 & 0.431 &
        & 77.96 & 92.23 & 0.375 &
        & 80.21 & 93.87 & 0.3378 &
        & 70.62 & 90.42 & 0.3926 \\

    CoKT
        & \second{79.62} & \second{91.18} & 0.3904 &
        & 79.25 & 87.84 & 0.396 &
        & 71.40 & 87.50 & 0.3794 &
        & 78.95 & 92.60 & \second{0.336} &
        & 73.09 & 91.32 & 0.3664 \\

    CoSKT
        & 77.23 & 87.86 & 0.3993 &
        & 73.18 & 81.62 & 0.4467 &
        & 78.66 & 92.26 & 0.3768 &
        & 77.92 & 93.32 & 0.3526 &
        & 58.96 & 82.07 & 0.4978 \\

    RACL
        & 79.55 & 91.17 & 0.3910 &
        & 81.26 & 88.05 & 0.403 &
        & \second{79.05} & \second{93.01} & \best{0.3708} &
        & \second{80.97} & 94.19 & 0.3373 &
        & 70.51 & 90.14 & 0.368 \\

    FINER
        & 78.30 & 90.44 & 0.3958 &
        & 79.31 & 86.65 & 0.4133 &
        & 78.97 & 92.45 & 0.3759 &
        & 76.32 & 92.38 & 0.3523 &
        & 67.99 & 89.58 & 0.3791 \\

    \hline

    \textbf{CaRe-KT}
        & \best{79.98} & \best{91.87} & \second{0.3901} &
        & \best{82.94} & \best{90.01} & \second{0.3788} &
        & \best{80.55} & \best{94.37} & \second{0.3714} &
        & \best{82.79} & \best{95.33} & \best{0.3285} &
        & \best{78.50} & \best{94.56} & \best{0.3565} \\

    \hline
    \end{tabular}}
\end{table*}
%
%
As shown in Table~\ref{OveralPerformance}, CaRe-KT achieves the best AUC and AP on all five datasets, improving AUC over the strongest baseline by 0.36 (DBE-KT22), 1.49 (ASSISTment09), 1.50 (Slepemapy.cz), 1.82 (Junyi) and 3.11 (XES3G5M), and AP by 0.69 to 2.16. The strongest baseline differs across datasets (CoKT, stableKT, RACL, RACL and DyGKT, respectively), suggesting that existing designs are tailored to particular data characteristics, whereas CaRe-KT generalizes across them. The largest gain occurs on XES3G5M, where only 11\% of test interactions involve questions seen during training. Question-specific parameters cannot be learned for unseen questions, which most severely affects CoSKT (58.96 AUC, below DKT), as its prediction layer is indexed by question identity. CaRe-KT instead represents questions by their content (Eq.~(17)), so the EGC can retrieve evidence for unseen questions through related ones. On Junyi and Slepemapy.cz, where large learner populations provide dense cross-learner evidence, the strongest baselines exploit cross-learner information (RACL, FINER, CoSKT), yet CaRe-KT still adds 1.82 and 1.50 AUC and attains the lowest RMSE on Junyi, consistent with the benefit of preserving acquisition relevance during reasoning. The margin is narrowest on DBE-KT22, where a small, fully observed item pool favours all strong models and the top four lie within 0.43 AUC, although the AP gain (+0.69) is nearly twice the AUC gain. In probability estimation, CaRe-KT obtains the lowest RMSE on Junyi and XES3G5M and the second lowest elsewhere, trailing the best baseline (IEKT, stableKT and RACL, respectively) by at most 0.0007. None of these is competitive on RMSE on other datasets, and CaRe-KT is the only model ranked in the top two on every dataset for every metric. Its advantage therefore lies mainly in discrimination, while its calibration matches the best dataset-specific baseline.
\subsubsection{RQ2. Robustness under Sparse and Cold-Start Conditions}

\paragraph{Cold-start}
Table~\ref{QuestionColdStart} reports performance on questions withheld from training. CaRe-KT achieves the best AUC and AP on all five datasets and the lowest RMSE on four, with RACL marginally ahead on DBE-KT22 (0.4070 vs.\ 0.4073). Its AUC margin over the strongest baseline widens relative to Table~\ref{OveralPerformance} on every dataset, reaching +1.18 on DBE-KT22, +2.46 on ASSISTment09, +4.71 on Slepemapy.cz, +3.19 on Junyi and +6.17 on XES3G5M (vs.\ +0.36 to +3.11). Models relying on question-specific parameters or per-question histories degrade sharply, as DyGKT loses 21.30 AUC on Junyi and 13.06 on Slepemapy.cz, CoKT loses 12.26 on DBE-KT22, and CoSKT falls below 60 AUC everywhere. RACL and FINER remain the two strongest baselines on ASSISTment09, Slepemapy.cz and Junyi, consistent with peer evidence staying accessible for withheld questions, yet RACL still loses 6.11 and 5.05 AUC on the latter two. CaRe-KT loses only 0.59--1.55 AUC on XES3G5M, Slepemapy.cz and ASSISTment09, as content-based compatibility (Eqs.~(4)-(5)) links a withheld question to related ones without its own response history. Its larger decreases on DBE-KT22 ($-3.17$) and Junyi ($-3.68$) coincide with the two smallest item pools (212 and 0.7k questions), which leave few candidates in $\mathcal{N}_\epsilon(q_t)$.
%

\begin{figure}
    \centering
    \includegraphics[width=1\linewidth]{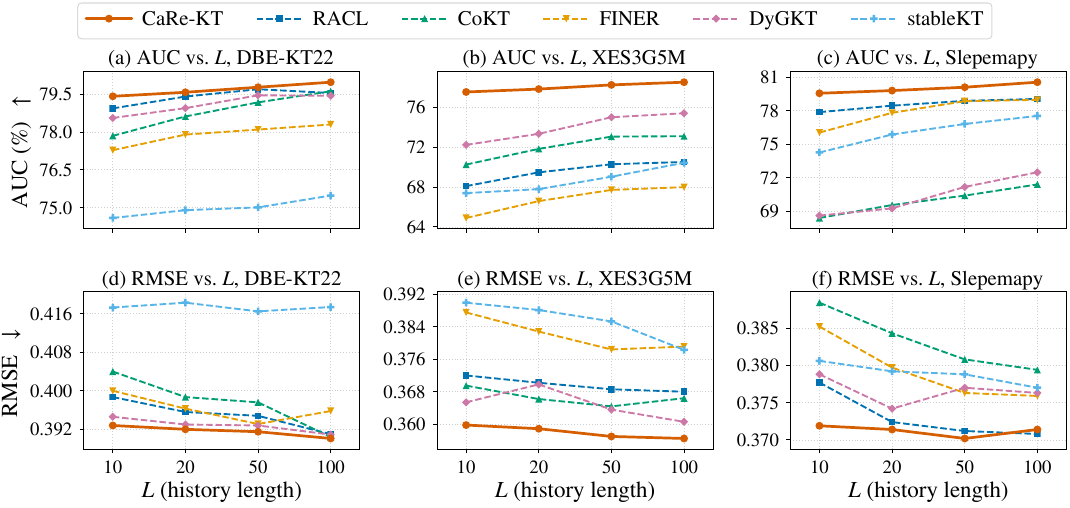}
         \caption{Performance under restricted learner history. AUC (top) and RMSE (bottom) of CaRe-KT and five baselines on DBE-KT22 (a, d), XES3G5M (b, e), and Slepemapy.cz (c, f) when each target learner's history is limited to its most recent $L$ interactions, with other learners' data unchanged.}
    \label{sparsity}
\end{figure}

\begin{table*}[!ht]
    \caption{Question-level cold-start performance on five datasets. The best result in each column is shaded dark blue and in bold, while the second best is shaded light blue and underlined. Arrows indicate the preferred direction ($\uparrow$ higher is better, $\downarrow$ lower is better)}
    \label{QuestionColdStart}
    \centering
    \scalebox{0.87}{\begin{tabular}{lccccccccccccccccccc}
    \hline
    \multirow{2}{*}{\textbf{Model}} & \multicolumn{3}{c}{\textbf{DBE-KT22}} & & \multicolumn{3}{c}{\textbf{ASSISTment09}} && \multicolumn{3}{c}{\textbf{Slepemapy.cz}} && \multicolumn{3}{c}{\textbf{Junyi}} && \multicolumn{3}{c}{\textbf{XES3G5M}} \\
    \cline{2-4} \cline{6-8} \cline{10-12}\cline{14-16} \cline{18-20}
        & $\uparrow$AUC & $\uparrow$AP & $\downarrow$RMSE &&  $\uparrow$AUC & $\uparrow$AP & $\downarrow$RMSE && $\uparrow$AUC & $\uparrow$AP & $\downarrow$RMSE && $\uparrow$AUC & $\uparrow$AP & $\downarrow$RMSE && $\uparrow$AUC & $\uparrow$AP & $\downarrow$RMSE \\
    \hline
    DKT
        & \second{75.63} & 85.71 & 0.4347 &
        & 58.67 & 69.99 & 0.478 &
        & 59.99 & 82.98 & 0.422 &
        & 57.33 & 81.93 & 0.483 &
        & 69.83 & 89.15 & 0.3894 \\

    AKT
        & 74.77 & 87.39 & 0.4474 &
        & 62.99 & 73.23 & 0.479 &
        & 59.34 & 83.09 & 0.440 &
        & 57.39 & 82.51 & 0.432 &
        & 70.52 & 89.08 & 0.3954 \\

    IEKT
        & 73.98 & 87.52 & 0.4447 &
        & 63.24 & 73.48 & 0.469 &
        & 62.22 & 83.94 & 0.418 &
        & 57.47 & 82.31 & 0.421 &
        & \second{71.74} & 89.88 & 0.3823 \\

    DIMKT
        & 59.05 & 81.69 & 0.4913 &
        & 64.84 & 74.30 & 0.460 &
        & 54.78 & 79.78 & 0.418 &
        & 53.08 & 80.28 & 0.409 &
        & 70.36 & 89.02 & 0.3909 \\

    QIKT
        & 70.76 & 82.71 & 0.4359 &
        & 66.24 & 73.50 & 0.474 &
        & 62.58 & 84.56 & 0.418 &
        & 58.56 & 83.03 & 0.406 &
        & 71.14 & 89.26 & 0.3886 \\

    simpleKT
        & 72.92 & 84.29 & 0.4401 &
        & 64.11 & 73.98 & 0.479 &
        & 63.96 & 85.31 & 0.409 &
        & 57.48 & 82.40 & 0.411 &
        & 69.69 & 88.68 & 0.3862 \\

    DyGKT
        & 69.72 & 86.92 & 0.4851 &
        & 60.23 & 71.02 & 0.478 &
        & 59.43 & 81.98 & 0.462 &
        & 59.05 & 83.42 & 0.407 &
        & 71.26 & \second{89.94} & 0.3849 \\

    stableKT
        & 72.63 & 86.72 & 0.4656 &
        & 71.84 & 80.91 & 0.442 &
        & 64.10 & 85.78 & 0.408 &
        & 69.62 & 88.54 & 0.385 &
        & 70.42 & 89.06 & 0.3854 \\

    UKT
        & 69.81 & 86.61 & 0.4270 &
        & 75.58 & 84.98 & 0.425 &
        & 66.44 & 87.14 & 0.403 &
        & 70.37 & 89.35 & 0.386 &
        & 70.15 & 88.94 & 0.3908 \\

    CoKT
        & 67.36 & 85.83 & 0.4168 &
        & 71.55 & 81.59 & 0.458 &
        & 63.52 & 84.31 & 0.453 &
        & 73.61 & 82.33 & 0.402 &
        & 65.81 & 87.94 & 0.4012 \\

    CoSKT
        & 57.00 & 79.69 & 0.499 &
        & 59.38 & 64.36 & 0.4808 &
        & 55.99 & 79.47 & 0.485 &
        & 58.77 & 86.00 & 0.496 &
        & 57.18 & 79.67 & 0.503 \\

    RACL
        & 72.96 & \second{89.18} & \best{0.407} &
        & 77.26 & 84.85 & 0.427 &
        & 72.94 & 90.69 & 0.3854 &
        & \second{75.92} & \second{93.13} & 0.346 &
        & 69.66 & 88.74 & 0.385 \\

    FINER
        & 71.65 & 88.54 & 0.409 &
        & \second{78.93} & \second{86.53} & \second{0.4117} &
        & \second{74.90} & \second{91.62} & \second{0.380} &
        & 75.74 & 92.75 & \second{0.3446} &
        & 70.73 & 89.37 & \second{0.382} \\

    \hline

    \textbf{CaRe-KT}
        & \best{76.81} & \best{90.77} & \second{0.4073} &
        & \best{81.39} & \best{88.79} & \best{0.4036} &
        & \best{79.61} & \best{93.54} & \best{0.3771} &
        & \best{79.11} & \best{94.45} & \best{0.3406} &
        & \best{77.91} & \best{94.22} & \best{0.3601} \\

    \hline
    \end{tabular}}
\end{table*}
\paragraph{Sparse history} To emulate learners with limited interaction histories, each target learner's history is restricted to its most recent $L\in\{10,20,50,100\}$ interactions, while other learners' data remain unchanged and all models are retrained for each $L$ (Fig.~\ref{sparsity}). The comparison spans three datasets of different scale and content type and includes every baseline that is strongest on at least one dataset in Table~\ref{OveralPerformance} (CoKT, stableKT, RACL, and DyGKT), together with FINER. Among these models, CaRe-KT achieves the best AUC at every $L$ on DBE-KT22, XES3G5M, and Slepemapy.cz. Its margin over the strongest baseline is largest at the shortest history on XES3G5M (5.30 at $L=10$ vs.\ 3.11 at $L=100$) and Slepemapy.cz (1.69 vs.\ 1.50), and remains narrow on DBE-KT22 (0.08-0.48), where the leading models perform comparably. Reducing the history from 100 to 10 interactions lowers the AUC of CaRe-KT by only 0.56-0.99, a smaller decrease than that of any baseline on the same dataset. On XES3G5M and Slepemapy.cz, CaRe-KT with only ten interactions outperforms every baseline with 100. This robustness is consistent with the EGC acquiring evidence from the query question and its temporal context, independently of the length of the learner's own history. CaRe-KT also attains the lowest RMSE in all settings except $L=100$ on Slepemapy.cz, where RACL is marginally lower.

\subsubsection{Ablation Study (RQ3-RQ5)}
\begin{table}[t]
\caption{Ablation results for CaRe-KT variants on DBE-KT22, XES3G5M, and Slepemapy.cz, evaluated by AUC and RMSE.}
\label{tab:ablation}
\centering\footnotesize\setlength{\tabcolsep}{2pt}
\scalebox{0.98}{\begin{tabular}{@{}l cc c cc c cc@{}}
\toprule
& \multicolumn{2}{c}{\textbf{DBE-KT22}} && \multicolumn{2}{c}{\textbf{XES3G5M}} && \multicolumn{2}{c}{\textbf{Slepemapy.cz}}\\
\cmidrule{2-3}\cmidrule{5-6}\cmidrule{8-9}
\textbf{Variant} & $\uparrow$AUC & $\downarrow$RMSE && $\uparrow$AUC & $\downarrow$RMSE && $\uparrow$AUC & $\downarrow$RMSE\\
\midrule
CaRe-KT (full) & \textbf{79.98} & \textbf{0.3901} && \textbf{78.50} & \textbf{0.3565} && \textbf{80.55} & \textbf{0.3714} \\
\midrule
\multicolumn{9}{@{}l}{\textit{\textbf{RQ3: relevance continuity}}}\vspace{0.1cm}\\
w/o Acq.\ Relevance & 79.68 & 0.3912 
                    && 76.55 & 0.3633
                    && 79.50 & 0.3748 \\
w/o Compatibility   & 79.80 & 0.3908 
                    && 77.75 & 0.3591 
                    && 79.85 & 0.3732 \\
Learned-$\gamma$    & 79.92 & 0.3902 
                    && 78.40 & 0.3569 
                    && 80.20 & 0.3722 \\
MLP Fusion          & 79.78 & 0.3908 
                    && 77.40 & 0.3602 
                    && 79.72 & 0.3735 \\
Post-hoc Reweight   & 79.72 & 0.3910 
                    && 76.90 & 0.3620 
                    && 79.60 & 0.3744 \\                  
\midrule
\multicolumn{9}{@{}l}{\textit{\textbf{RQ4: query-adaptive evidence}}}\vspace{0.1cm}\\
w/o Retrieval       & 79.11 & 0.3930 
                    && 74.56 & 0.3682  
                    && 72.83 & 0.3787 \\
Top-$K$ Evidence    & 79.70 & 0.3907 
                    && 77.55 & 0.3598 
                    && 79.75 & 0.3734 \\
w/o Temp.\ Decay    & 79.76 & 0.3905 
                    && 78.00 & 0.3583 
                    && 79.65 & 0.3737 \\
Semantic Only       & 79.72 & 0.3908 
                    && 78.15 & 0.3577 
                    && 79.65 & 0.3737 \\
Concept Only        & 79.78 & 0.3904 
                    && 76.50 & 0.3635 
                    && 79.95 & 0.3729 \\
%
\midrule
\multicolumn{9}{@{}l}{\textit{\textbf{RQ5: structured representation}}}\vspace{0.1cm}\\
w/o $\mathcal{L}_{\mathrm{SCL}}$  & 79.76 & 0.3909 
                                  && 77.90 & 0.3586 
                                  && 79.85 & 0.3732 \\
w/o $\mathcal{L}_{\mathrm{self}}$ & 79.81 & 0.3906 
                                  && 78.20 & 0.3576 
                                  && 80.00 & 0.3728 \\
w/o $\mathcal{L}_{\mathrm{reg}}$  & 79.70 & 0.3911 
                                 && 77.50 & 0.3600 
                                 && 79.75 & 0.3734 \\
\bottomrule
\end{tabular}}
\end{table}
To isolate the contribution of each component, Table~\ref{tab:ablation} removes or replaces one at a time under the protocol of Section~V-A4, grouping rows by the two terms of relevance-informed attention (RQ3), the presence and construction of EGC evidence (RQ4), and the two regularizers (RQ5). Since component contributions depend on the availability and characteristics of supporting evidence, we conduct the ablation on three datasets with three distinct evidence regimes.
On DBE-KT22, every test question has its own response history, and the three strongest models lie within 0.4 AUC (Table~III), providing a setting with strong question-specific evidence and limited performance separation. On XES3G5M, 89\% of test interactions involve questions unseen during training, making cross-question evidence acquisition particularly important. On Slepemapy.cz, 91k learners provide a large pool of historical interactions, making evidence selection and weighting especially relevant to examine. ASSISTment09 and Junyi are excluded owing to sparse evidence (Section~V-A1) and computational cost, respectively.
\paragraph{RQ3. Effect of relevance continuity}
Relevance continuity is realized through the two terms of Eq.~(26). Removing the acquisition relevance $\mathcal{W}^E_t$, so that attention rests on learned compatibility alone, lowers AUC by 0.30, 1.95 and 1.05 on DBE-KT22, XES3G5M and Slepemapy.cz, with RMSE rising in step, the largest loss among the weighting variants on every dataset. On DBE-KT22 this single change reduces the margin over CoKT from 0.36 to 0.06 AUC. Removing the learned compatibility $e_{ti}$ instead costs 0.18, 0.75 and 0.70, so acquisition relevance makes the larger contribution while learned compatibility adapts the weighting to the query, and neither term alone recovers the full model. Three alternative fusions test whether the additive form itself is responsible. Replacing the unit weight on $\mathcal{W}^E_t$ with a learned coefficient (Learned-$\gamma$) costs only 0.06, 0.10 and 0.35, with $\gamma$ converging to 0.9, 1.3 and 1.1. Supplying $\mathcal{W}^E_t$ as an input to a one-hidden-layer compatibility function (MLP Fusion) costs 0.20, 1.10 and 0.83, and training without it and rescaling $\alpha_{ti}$ by it at inference (Post-hoc Reweight), as a retrieve-then-predict pipeline would, costs 0.26, 1.60 and 0.95, close to removing the term entirely. The gain therefore comes from relevance entering reasoning as a prior alongside, rather than inside, the learned compatibility, and from attention being trained in its presence.
\paragraph{RQ4. Effectiveness of query-adaptive evidence construction}
The contribution of the EGC is assessed first by removing the evidence it constructs and then by altering how it is constructed. Removing retrieval altogether, while keeping the content-based question representations of Eq.~(17), costs 0.87, 3.94 and 7.72 AUC on DBE-KT22, XES3G5M and Slepemapy.cz, the largest loss in Table~V on every dataset. The gain on unseen questions is therefore not explained by semantic embeddings alone, and the loss grows with the density of cross-learner evidence. Among the construction choices, a fixed budget equal to the mean anchor count in place of the adaptive boundary $\epsilon_t$ costs 0.28, 0.95 and 0.80 AUC, with RMSE rising in step, so a uniform evidence cardinality is inferior to a query-specific one even when the average number of anchors is matched. Removing temporal decay costs 0.22, 0.50 and 0.90, the largest construction loss on Slepemapy.cz, indicating that distinguishing historical occurrences by their temporal validity contributes beyond question-level relevance. The two compatibility signals of Eq.~(4) differ in importance across datasets. On XES3G5M, whose questions carry text and are mostly unseen at test time, concept overlap alone loses 2.00 AUC, the largest change among the construction variants and marginally more than removing acquisition relevance, whereas semantic similarity alone loses 0.35. On Slepemapy.cz, whose content is metadata, the order reverses (0.60 against 0.90), supporting the combination of both signals rather than reliance on either alone.
\paragraph{RQ5. Effect of behaviorally structured representation learning}
The representation space over which attention operates is shaped by $\mathcal{L}_{\mathrm{SCL}}$ on the retrieved trajectories and $\mathcal{L}_{\mathrm{self}}$ on the target history. Removing both costs more than removing either. On DBE-KT22 the loss is 0.28 AUC against 0.22 for $\mathcal{L}_{\mathrm{SCL}}$ and 0.17 for $\mathcal{L}_{\mathrm{self}}$, on XES3G5M 1.00 against 0.60 and 0.30, and on Slepemapy.cz 0.80 against 0.70 and 0.55. The contrastive term is the larger of the two throughout, consistent with its role in shaping the evidence space that attention reads, while the consistency term provides an additional, smaller gain. Together, the regularizers improve prediction by shaping complementary aspects of the behavioral representations used in relevance-informed reasoning.
%
\subsubsection{Sensitivity Analysis}
\begin{figure}
    \centering
    \includegraphics[width=1\linewidth]{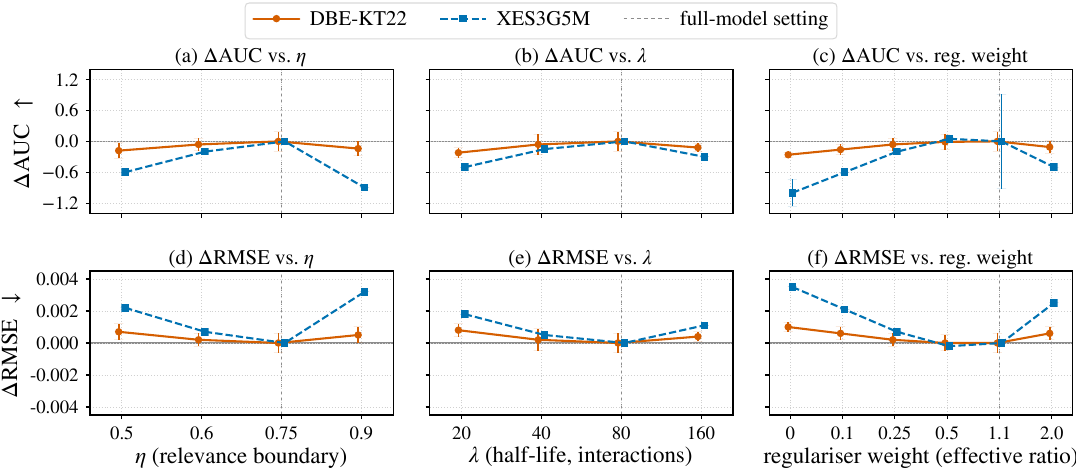}
    \caption{Sensitivity of CaRe-KT to (a, d)~$\eta$, (b, e)~$\lambda$, and (c, f)~the regularization weight, as the change in AUC $\uparrow$  and RMSE $\downarrow$ relative to the full model of Table~\ref{tab:ablation}, marked by the dashed lines. Mean $\pm$ one standard deviation over three seeds on DBE-KT22 and one seed on XES3G5M. The regularization axis is the effective ratio of $\mathcal{L}_{\mathrm{reg}}$ to $\mathcal{L}_{\mathrm{pred}}$.}
    \label{fig:sens}
\end{figure}

We examine how far the performance of CaRe-KT changes when each of the three hyperparameters that control its mechanisms is varied from the configuration of Section~V-A4 while all other settings are held fixed. The relevance boundary $\eta$ (Eq.~\ref{adaptivebound}) is varied over $\{0.5, 0.6, 0.75, 0.9\}$, the decay half-life $\lambda$ (Eq.~\ref{eq:dynamic_relevance}) over $\{20, 40, 80, 160\}$ interactions, and the weight of $\mathcal{L}_{\mathrm{reg}}$, reported as the effective ratio of the regularization term to the prediction loss (Section~V-A4), over $\{0, 0.1, 0.25, 0.5, 1.1, 2.0\}$, where $0$ is the variant without $\mathcal{L}_{\mathrm{reg}}$ in Table~\ref{tab:ablation} and $1.1$ is the full model. Each point is retrained from scratch and selected on validation loss as in the main experiments. Fig.~\ref{fig:sens} reports the resulting change in test AUC and RMSE relative to the full model on DBE-KT22 (three seeds) and XES3G5M, two datasets that differ by an order of magnitude in interactions and in the share of unseen test questions.

On DBE-KT22, no setting departs from the default by more than the seed-to-seed variation of the full model (standard deviation 0.19 AUC over three seeds), so on this dense and fully observed item pool none of the three parameters requires tuning. On XES3G5M, where 89\% of test questions are unseen, each curve peaks at or near the default and falls at both ends. Tightening the boundary ($\eta{=}0.9$) costs more than relaxing it (0.9 against 0.6 AUC), because an unseen question must draw its evidence from related ones. A short half-life excludes older occurrences that are still informative, whereas a long one approaches the no-decay variant of Table~\ref{tab:ablation}. The regularization weight shows a plateau between 0.25 and 1.1, and at 2.0 RMSE degrades faster than AUC. The defaults therefore lie within a stable region on both datasets.
%
%
\section{Discussion}

\subsection{Principal Findings}
The results show that CaRe-KT is most effective when target-specific information is scarce. Its margin is largest on XES3G5M, where most test questions are unseen, it widens on all five datasets when questions are withheld from training, and it is largest at the shortest learner histories on XES3G5M and Slepemapy.cz.
Conventional KT models are weakest in these conditions, as their estimates rest on parameters and histories tied to the target learner or question. CaRe-KT instead draws on evidence acquired through question content and temporal validity, and retains the relevance of that evidence when weighting it. The ablation attributes this advantage to the design as a whole rather than to any single part of it. Of the two terms of relevance-informed attention, acquisition relevance is the one whose removal costs more on every dataset. The query-adaptive EGC matters most where evidence must be drawn from related questions, as on XES3G5M, and the two regularizers contribute a smaller but consistent gain through the representations that attention reads. The framework also has limits. The margin is narrow where strong models converge on a small, fully observed item pool, as on DBE-KT22, and the gains concern ranking more than probability error.
\subsection{Educational Implications}
In learning technologies, a knowledge estimate matters through the decisions it informs, such as which exercise to recommend, when to advance a learner, and where an educator should intervene. The findings bear on four aspects of such decisions.

\textit{Early adaptivity for learners with short records.} In a learner's first interactions, adaptive systems often rely on default sequences or population-level estimates~\cite{shen2024survey, long2022improving}, so the learners about whom least is known receive the least personalized support.
CaRe-KT loses at most 0.99 AUC points when the visible history is reduced from 100 to 10 interactions (Section~V-B), and on XES3G5M and Slepemapy.cz its predictions from ten interactions are more accurate than those of every compared baseline given 100. This suggests that remediation and progression decisions could be personalized from the early stages of engagement.

\textit{Evidence-grounded estimates for educators.} A mastery estimate is a claim about a learner, and principled assessment requires such claims to be warranted by identifiable evidence~\cite{mislevy2003focus}. CaRe-KT attaches to each prediction the prior interactions it retrieved, the relevance that led to their retrieval, and the weight each received. Teacher dashboards and open learner models~\cite{bull2020there} could therefore present the comparable learning experiences behind a prediction alongside the prediction itself, giving educators grounds to judge whether similar behavior is in fact relevant evidence before acting on it.

\textit{ Tracing newly authored content.} On platforms where teachers author and share exercises, the question bank is continually revised~\cite{heffernan2014assistments}. Models that learn a parameter per question cannot trace a new item until responses to it accumulate~\cite{lu2022cmkt, zhang2024coskt}. CaRe-KT represents a question by its content (Eq. 17) and links it through the EGC to related questions. A new item is therefore supported by the trajectories of learners who practiced related material from the time it is first assigned. Under question-level cold start, CaRe-KT remains the most accurate model on all five datasets, including the three that provide only metadata (Table IV). Since the cache and memory are updated online (Section IV-F), teachers could add or revise exercises during a term without retraining the model. This advantage narrows on small question banks, which contain few related items (Section VI-C).

\textit{Matching estimates to instructional decisions.} A knowledge estimate serves two kinds of instructional decision. Some decisions order learners or exercises relative to one another, whereas others compare a predicted probability with a fixed criterion~\cite{corbett1994kt}. CaRe-KT achieves the best AUC and AP on all five datasets, and its RMSE is the lowest on two datasets in the overall evaluation (Table~III), on four under question-level cold start (Table~IV), and in all but one sparse-history setting (Fig.~3). Its advantage is therefore most consistent in ranking correct responses above incorrect ones, which makes its predictions well suited to decisions of the first kind, such as which learners need attention first or which exercise to present next. For decisions of the second kind, the predicted probabilities should be calibrated and verified before deployment, which is a standard post-hoc procedure~\cite{qian2025extending} rather than an obstacle to such use.

\subsection{Limitations and Future Work}
Three considerations define the scope of this study and the work that follows from it. First, in line with standard practice in KT, the evaluation is offline and measures the prediction of logged responses. Whether more reliable early estimates lead to better instructional decisions and learning outcomes is a question for classroom deployment, where the evidence subgraph can also be developed into teacher-facing explanations through co-design with educators. Second, CaRe-KT draws its strength from question content. It remained effective with metadata descriptions alone, but its cold-start losses were larger on the two smallest item pools, which suggests that richer content encoders would benefit domains with few or sparsely described questions. Third, CaRe-KT is designed for online use, in which the interactions of other learners become available as they occur, and our evaluation follows this setting. Deployments in which such data are delayed or cannot be shared would have less evidence to draw on, which makes privacy-preserving retrieval of learner trajectories an important direction.
\section{Conclusion}

This paper identified a retrieval–inference alignment gap in retrieval-augmented KT, in which relevance governs what evidence is retrieved but is not preserved when its influence is determined. We proposed relevance continuity as a principle for closing this gap and realized it in CaRe-KT, which acquires temporally valid evidence through the Evidence-Graph Cache, encodes it in a behaviorally structured space, and carries acquisition relevance into reasoning as a prior that informs without dictating the influence of evidence. Across five benchmarks, CaRe-KT achieved the best ranking performance against thirteen baselines, and its advantage widened on all datasets under cold-start and sparse-history conditions, where reliable knowledge estimation is hardest. Ablation studies on three benchmarks showed that removing acquisition relevance consistently produced the largest performance loss among the attention variants, a loss that post-hoc reweighting at inference mitigated only marginally. These findings provide a foundation for adaptive learning systems whose estimates remain reliable and traceable to their evidence when learner data are scarce.
%
\section*{Acknowledgments}
This research was supported by the Australian Government throughthe Australian Research Council’s Industrial Transformation Training Centre for Information Resilience (CIRES) project number IC200100022.
\bibliographystyle{IEEEtran}
\bibliography{reference}

\end{document}